\documentclass[a4paper]{article}

\usepackage{INTERSPEECH2022}
\usepackage{subfiles}
\usepackage{cleveref}
\usepackage{todonotes}
\usepackage{dblfloatfix} 
\usepackage{amsmath,graphicx}
\usepackage{multirow}
\usepackage{subfigure}
\usepackage{tabularx}
\usepackage[hyphens]{url}

\newcommand{\datasetname}{Earnings-22}
\newcommand{\github}{\url{https://github.com/revdotcom/speech-datasets/tree/master/earnings22}}

\title{\datasetname: A Practical Benchmark for Accents in the Wild}

\name{
Miguel Del Rio,
Peter Ha,
Quinten McNamara,
Corey Miller,
Shipra Chandra
}

\address{
Rev.com
}
\email{
miguel.delrio@rev.com
}

\begin{document}

\maketitle

\begin{abstract}
Modern automatic speech recognition (ASR) systems have achieved superhuman Word Error Rate (WER) on many common corpora despite lacking adequate performance on speech in the wild. Beyond that, there is a lack of real-world, accented corpora to properly benchmark academic and commercial models. To ensure this type of speech is represented in ASR benchmarking, we present Earnings-22, a 125 file, 119 hour corpus of English-language earnings calls gathered from global companies. We run a comparison across 4 commercial models showing the variation in performance when taking country of origin into consideration. Looking at hypothesis transcriptions, we explore errors common to all ASR systems tested. By examining Individual Word Error Rate (IWER), we find that key speech features impact model performance more for certain accents than others. Earnings-22 provides a free-to-use benchmark of real-world, accented audio to bridge academic and industrial research.
\end{abstract}


\section{Introduction}
Automatic speech recognition systems are utilized in domains ranging from finance to media by providing a rich transcription of the original audio. However, these systems are often error-prone when tasked with transcribing audio with varied features, such as accents, noise, and unique vocal characteristics. Part of the reason for this inaccuracy is that the types of data that are used to train and prepare these models are not always indicative of their use case. For example, an ASR model trained exclusively on American-accented speech using a high-quality microphone is likely to perform poorly if used to transcribe speech from a person speaking Indian-accented English. One way to help identify these issues is to evaluate them using metrics such as WER, IWER\cite{goldwater2010words}, and other word characteristics to examine the specific ways in which the models fail.

Evaluation datasets which provide a variety of accented speech, along with high quality reference text, are often difficult to obtain. These difficulties arise from having multilingual speakers, non-English disfluencies, and accented speech, to name a few. Company earnings calls are a great source of this type of data, as they provide a large amount of speech variety from having many different speakers, differing accents, and complex domain terminology. Although they contain a concentration of financial jargon, earnings calls still provide a broad coverage of real-world topics. In this paper, we present a compiled dataset of 119 hours that covers 7 regional accents and is freely available to the public. We perform WER analysis using several industry ASR models and compare their performance across accent regions and word characteristics. In \Cref{dataset_description_section}, we describe the data properties and collection methodology. In \Cref{wer_analysis_section} and \Cref{model_bias_section}, we provide an initial analysis of accuracy disparities between regional accents. Finally, we conclude with a call to action to promote improved accent bias benchmarking in the ASR field.

\section{The \textit{\datasetname} Dataset} \label{dataset_description_section}
The \textit{\datasetname} benchmark dataset\footnote{This benchmark is available on Github at \github.} is developed with the intention of providing real-world audio focused on identifying bias in ASR systems. Our attention focused on aggregation of accented public\footnote{Earnings calls fair use legal precedent in Swatch Group Management Services Ltd. v. Bloomberg L.P.} English-language earnings calls from global companies. We collected a total of 125 earnings calls, totalling 119 hours downloaded from various sources\footnote{Most calls are from \url{https://seekingalpha.com/}. Few come directly from the company websites: \url{https://www.mtn.com.gh/} and \url{https://transcorpnigeria.com/}}. 
The earnings calls in the \textit{\datasetname} corpus are sourced from a total of 27 unique countries which we categorize into 7 regions defined in \Cref{dataset:mapping}.
\begin{table}[h!]
    \begin{tabular}{l p{40mm} }
        \centering
        \textbf{African}& Nigeria, Ghana\\
        \textbf{Asian}& Indonesia, Turkey, India, Japan, South Korea, China\\
        \textbf{English}& United Kingdom, Canada, Australia, United States, South Africa\\
        \textbf{Germanic}& Denmark, Sweden, Germany\\
        \textbf{Other Romance}& France, Italy, Greece\\
        \textbf{Slavic}& Russia, Poland\\
        \textbf{Spanish / Portuguese}& Argentina, Brazil, Chile, Spain, Colombia, Portugal\\
    \end{tabular}
    \caption{Countries included in each language region. See \Cref{defining_regions} for further explanation on how these regions were defined.}
    \label{dataset:mapping}
\end{table}

To produce a broad range of speakers and accents, we focused our efforts on finding unique earnings calls from global companies. The process of properly labeling speaker accents is a difficult task that requires language experts to rate accents and techniques to deal with any disagreements in addition to implicit bias we add as a result of the rating. We opted to instead follow the method used in \cite{oneill21_interspeech} and associate an earnings call with the country of origin where the company was headquartered. 

To ensure diversity in the call selection, we opted to randomly select 5 calls from as many countries as were available to us. The only exception to this were calls from Ghana and Nigeria, which we actively sought out to improve the coverage of African accents in this dataset. Despite best efforts for these accents, we were only able to find 1 Nigerian and 4 Ghanaian earnings calls. Although these countries were the least represented in the dataset, their inclusion was crucial to improve the overall analytical value of the corpus.

\subsection{Creating and Preparing the Transcripts}
To ensure high quality transcripts, we submitted our files to our own human transcription platform. Once completed, the quality of each transcript is also verified by a separate group of graders. Following our work in \cite{delrio21_interspeech}, we chose to produce verbatim\footnote{For more information on Rev.com’s verbatim transcription see \url{https://www.rev.com/blog/resources/verbatim-transcription}} transcriptions to best model real human speech. We processed the transcripts produced by Rev.com and removed atmospheric information\footnote{Examples include information about background music, coughs, and other non-speech noises} using our internal processing tools. These files are then converted into our $.nlp$ file format that tokenizes the transcript and contains metadata tagged by our Named Entity Recognition (NER) system.

\subsection{Defining Regions}\label{defining_regions}
Due to the large number of countries but varying range of representation, the regions we defined aggregated several countries to make bias analysis more practical and conclusive (we show some statistics of those regions in \Cref{dataset:time_dist}). The primary region grouping we use is defined with a mixture of language family features and geographical location.
Due to the overwhelmingly large amount of Spanish-speaking countries in the dataset compared to other Romance languages, we felt the separation of Spanish / Portuguese and Other Romance was necessary to get a better view on the accents. Portuguese was grouped in with Spanish because of its linguistic similarity as both are a part of the Ibero-Romance group of Romance languages\cite{spanish_portuguese}.
Previous work has found that despite the a more distant genetic relationship between Greek and Romance languages, their proximity and long term contact has resulted in significant association both lexically and phonologically\cite{greekincontact} -- as a result, we felt they best fit in the Other Romance region. 
We chose to split South African earnings calls from the African region to follow the distinctions made in previous works \cite{accents_of_english, van_rooy_2020} denoting South African English as more resemblant of other countries in the English region than those in the African region\footnote{A resource comparing different accents, maintained by the author of \cite{accents_of_english} can be found at \url{https://www.phon.ucl.ac.uk/home/wells/accentsanddialects/}}.

\begin{table}[h!]
    \centering
    \begin{tabular}{|c|c|c|}
        \hline
        Language Regions & Time (in Hours) &  Number of Files \\
        \hline
        English&22.85&26\\
        Asian&25.27&28\\
        Slavic&7.72&10\\
        Germanic&13.53&12\\
        Spanish/Portuguese&28.87&31\\
        Other Romance&15.61&13\\
        African&5.06&5\\
        \hline
    \end{tabular}
    \caption{The \textit{\datasetname} corpus summarized by our defined linguistic regions, composed by considering region and language family.}
    \label{dataset:time_dist}
\end{table}

\section{WER Analysis on \textit{\datasetname}}
\label{wer_analysis_section}
To fully showcase the dataset and its characteristics, we provide an initial benchmark along the accent dimension. 
We used four cloud ASR providers to submit our evaluation audio and obtain hypothesis transcripts. For the Rev.ai models, there were two tested: a Kaldi based model and an end-to-end model.

\subsection{Provider and Regional File Breakdown}\label{weranalysis}
\begin{figure}[h]
\hspace*{-0.3cm}
\includegraphics[scale=0.4, width=8.45cm]{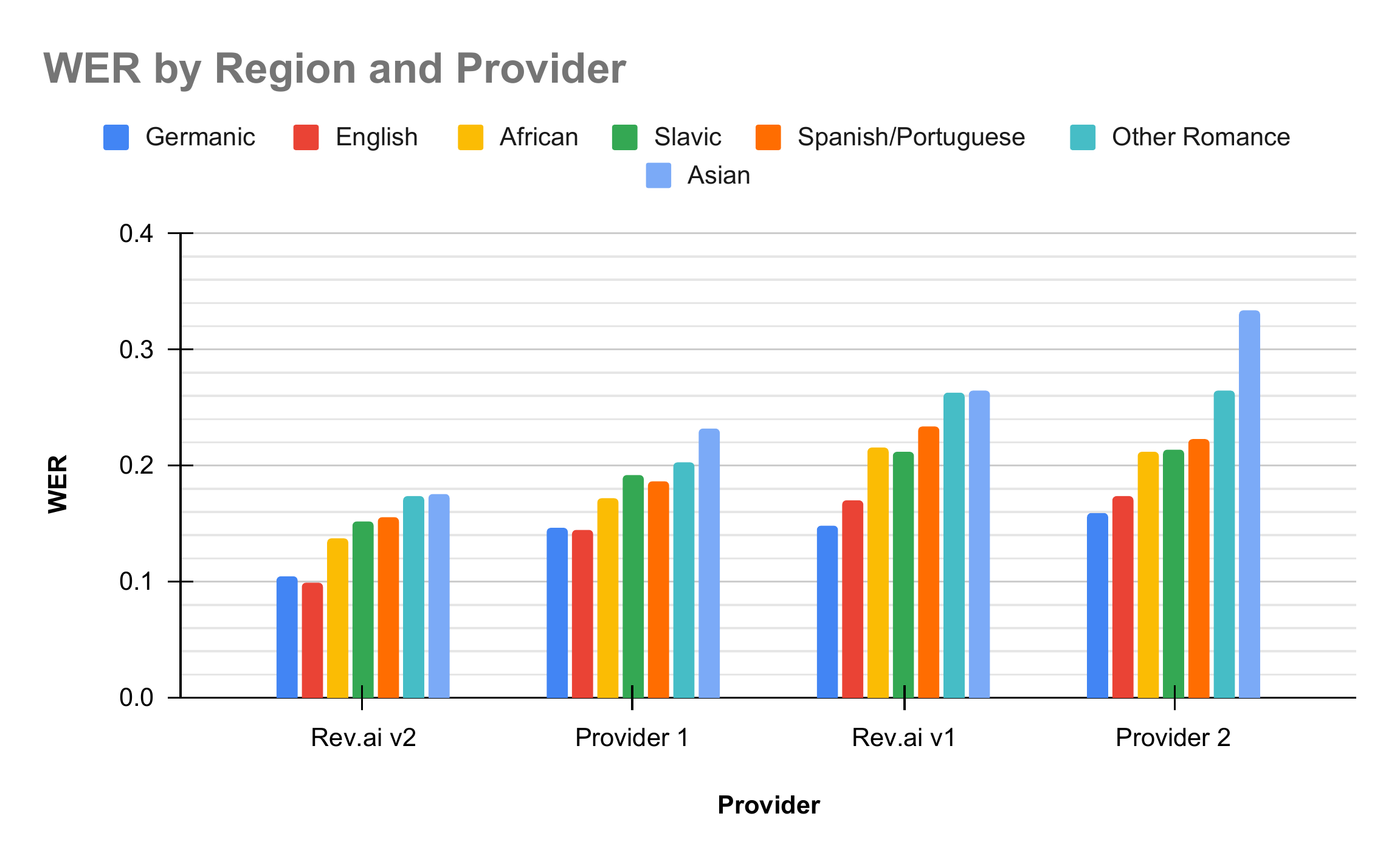}
\caption{WER by Region and Provider}
\label{wer_bar_graph}
\end{figure}

\Cref{wer_bar_graph} displays the average WER for each provider across accent regions. Here we aggregate by micro-averaging such that long files are weighted more than shorter files.

These results show that (1) English and Germanic regions seemingly perform the best with ASR relative to the other language regions and (2) Asian and Romance Languages other than Spanish and Portuguese perform the worst with ASR, relative to the other language regions. For English and Germanic, this distinction makes sense, as English is a Germanic language, and most modern ASR models are trained on English-region data. We speculate that the excellent performance on German may be due to highly skilled second-language speakers who may articulate more clearly and speak more slowly than first language speakers, a result also noted in \cite{miller2021corpus}.
The results on other regions echo the linguistic distance results of \cite{chiswick2005linguistic}. For example, the Asian region has the biggest WER gap from the English region. This is reinforced by the word level analysis in \Cref{regional_word_level_breakdown}. Moreover, we observe statistically significant differences in \Cref{model_bias_section}.

\subsection{Regional Word-Level Breakdown}\label{regional_word_level_breakdown}


\begin{table}[h]
\centering
\begin{tabular}{c}
\centering
\begin{tabular}{|c|}
\hline
Errors (All Files)\\
\hline
\color[HTML]{A569BD}{(capex, 27)}       \\
\color[HTML]{38761D}{(we'll, 17)}      \\
(yeah, 16)                               \\
\color[HTML]{A569BD}{ 
(ebitda, 12) }                           \\
\color[HTML]{38761D}{
(it's, 10)}                              \\
(an, 9)                                 \\
(too, 8)                                \\
(sales, 8)                              \\
\color[HTML]{38761D}{
(there's, 8)} \\
(cd, 8) \\
\hline
\end{tabular}%
\end{tabular}
\captionsetup{justification=centering}
\caption{Word-level errors across all files. Format of cell is (`word', number of files occurred in)\\
\color[HTML]{38761D} Contraction \color[HTML]{000000}$\vert$ \color[HTML]{A569BD} Financial Jargon}
\end{table}

We performed two methods of exploratory word-level error analysis. The first method of analysis involved combining word-level results for each file and provider separately and analyzing the words at the most granular level. The criteria for a word in this analysis was an F1 score of less than 0.3, and an occurrence within the transcript of five times or more. With these criteria, we sought to identify the most difficult words for the ASR systems. From this, we found that a couple domain-specific finance words were commonly incorrect, as well as a few contractions, whose processing often causes trouble for ASR systems\cite{goldwater2010words}. The results of this analysis are shown in Table 3.

Finally, we performed a word-level analysis broken down by region. We use the same F1 and frequency criteria as above, but also removed any words that met the criteria in every region. By doing so, we hoped to identify words that ASR systems struggled with particular to each region. The results of this analysis are shown in Table 4.

\begin{table*}[]        
\centering
\resizebox{\textwidth}{!}{%
\begin{tabular}{|l|l|l|l|l|l|l|}
\hline
\color[HTML]{000000} {African} &
  \color[HTML]{000000} {Asian} &
  \color[HTML]{000000} {English} &
  \color[HTML]{000000} {Germanic} &
  \color[HTML]{000000} {Other Romance} &
  \color[HTML]{000000} {Slavic} &
  \color[HTML]{000000} {Spanish/Portuguese} \\ \hline
\color[HTML]{A61C00} {(momo, 6)} &
  (an, 7) &
  \color[HTML]{38761D} {(we'll, 7)} &
  \color[HTML]{38761D} {(there's, 4)} &
  (euro, 5) &
  \color[HTML]{A61C00} {(cd, 8)} &
  \color[HTML]{BF9000} {(eh, 9)} \\
 (affect, 4) &
  (quarters, 6) &
  (stock, 5) &
  \color[HTML]{A61C00} {(reinhard, 3)} &
  (engineers, 3) &
  \color[HTML]{A61C00} {(projekt, 5)} &
  (yeah, 8) \\
\color[HTML]{3D85C6} {(gh, 2)} &
  (part, 4) &
  \color[HTML]{3D85C6} {(rand, 4)} &
  (federal, 3) &
  (grew, 3) &
  \color[HTML]{A61C00} {(cyberpunk, 3)} &
  \color[HTML]{A61C00} {(alejandro, 6)} \\
\color[HTML]{38761D} {(won't, 2)} &
  (there, 4) &
  (yeah, 4) &
  \color[HTML]{A61C00} {(lars, 3)} &
  \color[HTML]{A61C00} {(carlo, 3)} &
  (red, 3) &
  \color[HTML]{A61C00} {(chile, 6)} \\
 (better, 2) &
  \color[HTML]{3D85C6} {(tl, 4)} &
  \color[HTML]{A61C00} {(gus, 3)} &
  \color[HTML]{A61C00}{(aker, 3)} &
  (renovation, 3) &
  (rubles, 3) &
  \color[HTML]{38761D} {(it's, 6)} \\
\color[HTML]{38761D} {(we'll, 2)} &
  \color[HTML]{A61C00} {(won, 4)} &
  \color[HTML]{38761D} {(we're, 3)} &
  (lead, 3) &
  \color[HTML]{3D85C6} {(d, 3)} &
  (continued, 3) &
  (too, 5) \\
\color[HTML]{3D85C6} {(mtn, 1)} &
  \color[HTML]{3D85C6} {(mr, 4)} &
  (day, 3) &
  \color[HTML]{A61C00} {(roland, 3)} &
  \color[HTML]{38761D} {(we'll, 3)} &
  \color[HTML]{BF9000} {(hmm, 2)} &
  (had, 5) \\
(their, 1) &
  (business, 3) &
  (breakeven, 3) &
  \color[HTML]{A61C00} {(coke, 3)} &
  (asset, 3) &
  (higher, 2) &
  (euros, 5) \\
 (merchant, 1) &
  (monetization, 3) &
  \color[HTML]{A61C00} {(allen, 3)} &
  (billing, 3) &
  \color[HTML]{A61C00} {(tarek, 3)} &
  \color[HTML]{38761D} {(they're, 2)} &
  (slide, 5) \\
\color[HTML]{BF9000} {(oh, 1)} &
  (provisions, 3) &
  \color[HTML]{A61C00} {(billie, 3)} &
  (sea, 3) &
  \color[HTML]{A61C00} {(publicis, 2)} &
  \color[HTML]{3D85C6}{(gen, 2)} &
  (these, 5) \\ \hline
\end{tabular}%
}
\captionsetup{justification=centering}
\caption{Word-level errors by region and file. Format of cell is (`word', number of files occurred in)\\
\color[HTML]{A61C00} Name \color[HTML]{000000}$\vert$ \color[HTML]{38761D} Contraction \color[HTML]{000000}$\vert$ \color[HTML]{BF9000} Disfluency \color[HTML]{000000}$\vert$ \color[HTML]{3D85C6} Abbreviation/Acronym}
\end{table*}

In this table, we see that the most frequent errors in the Asian and Other Romance regions are common words as opposed to names, abbreviations, or acronyms. In the other regions, we see ASR systems struggle most on these kinds of specific terminology. This finding hints that the degradation of WER seen across region groups is not simply due to more industry-specific jargon or terminology, but poor recognition around common words spoken in regional accents.



\section{Exploring Model Bias}
\label{model_bias_section}
In \Cref{weranalysis}, we found that all providers are impacted when we consider the language region. Although these results indicate clear \textit{average} WER differences, we don't know for certain that these results are \textit{significant} as opposed to natural variations between files and models. In this section we run statistical tests and demonstrate model bias. In particular, we want to understand: (1) is the difference in WER statistically significant and (2) are there features of the transcripts that are impacting the model more. 
Though previously we've been considering multiple different models -- we focus in our statistical efforts on our newest Rev.ai V2 model. This is done because its the best performing of those we tested and we know the most details about the pipeline used to train it. 

\subsection{Measuring Significance}\label{montecarlo}
In the following experiments, we use a Monte Carlo Permutation Test\cite{10.5555/1196379} to evaluate our hypotheses. In our implementation, we compare two groups $A$ and $B$. After selecting some metric, M, to evaluate those groups, we define 
\begin{equation}
    \Delta = |\text{M}(A) - \text{M}(B)|
\end{equation}
as the absolute difference in between the metric evaluated on the groups.
We generate $n$ samples, $\mathbb{S}$, such that for the $i^{th}$ sample, we define the subset $A_i$ a random permutation of size $|A|$ from the set $A \bigcup B$ and define $B_i$ as $(A \bigcup B) \text{\textbackslash} A_i$. The $i^{th}$ sample is then 
\begin{equation}
    \delta_i = |\text{M}(A_i) - \text{M}(B_i)|
\end{equation}
After generating $n$ samples, the significance level is defined as the proportion of the samples that have a equal to or larger difference than $\Delta$.
\begin{equation}
    \text{p-value} = \frac{|\{\delta_j \geq \Delta \forall \delta_j \in \mathbb{S}\}|}{n}
\end{equation}

\subsection{Model Regional Bias}
As we see in \Cref{wer_bar_graph}, the Rev.ai V2 model performs differently in every language region. Since we know this model is trained on countries that predominantly belong to the English region, we expect that the differences we've observed should result in low p-values if the model is truly biased. In this section, we'll compare each region to the English region -- if any is found to be statistically different, this would imply that the model is significantly impacted by the defined region. 

\subsubsection{Experimental Setup}
We follow the experimental setup defined in \Cref{montecarlo}, and define $A$ to be the files in our baseline English region while $B$ is the region we want to evaluate. Our metric, M, for these experiments is the WER over the whole region. Unique to this setup is that instead of generating different permutations of the files, we took a more restrictive approach and generated combinations of files to ensure that each sample was unique\footnote{We also ran tests using permutations but found that the results are essentially identical as the number of samples increases.}. Finally, we generate 100,000 samples for each experiment and report the p-values in \Cref{bias:tab:file-permute}.

\begin{table}[h]
    \centering
    \begin{tabular}{|c|l|}
        \hline
         Region & p-value \\
        \hline
        African & 0.264\\
         Asian & 0.004$^{**}$\\
         Germanic & 0.928\\
         Other Romance & 0.012$^{*}$\\
         Slavic & 0.148\\
         Spanish / Portuguese & 0.035$^{*}$\\
         \hline
    \end{tabular}
    \caption{Monte Carlo Permutation Test results comparing the English Region to every other region we've defined. In the table above, p-values with a $^{*}$ are statistically significant at the 0.05 level while those with $^{**}$ are significant at the 0.005 level.}
    \label{bias:tab:file-permute}
\end{table}

\subsubsection{Results and Discussion}
We find that the Other Romance and Spanish / Portuguese region are statistically different from the English region at the 0.05 level while the Asian region is statistically different from the English region at the 0.005 level. 
These results reaffirm our findings in \Cref{weranalysis} showing that the three worst regions are statistically different from the English region. 
As previously mentioned, the work in \cite{oneill21_interspeech} found that assuming accents from corporate headquarters seemed to agree with a manual verification of a random sample of earnings calls -- given our results and previous work, we believe that these regional groupings \textit{do} reflect the presence of accented speech that causes our model to struggle.

\subsection{Transcript Features}
At Rev, we're particularly confident of our ASR model's ability to provide verbatim transcripts -- in particular, we know that our model is capable of capturing disfluencies like filled pauses and word fragments like false-starts. These words are notably harder than the average word to capture due to their relative infrequency but diverse methods of production. That being said, the work in \cite{goldwater2010words} showed that filled pauses did not have any impact on the recognition of surrounding words while fragmented words do impact a model when comparing the IWER between groups. We explore these features, conditioning on the linguistic differences defined by the various Language Regions.

\subsubsection{Experimental Setup}
We once again apply the same experimental setup described in \Cref{montecarlo}. Group $A$ is defined as the key words to test while group $B$ is all other words in the region. Our metric, M, for these experiments is IWER over the different groups. 
To define the groups in these experiments, we take advantage of Rev.com's style guide rules that provide fixed structure to key transcript features\footnote{\url{https://cf-public.rev.com/styleguide/transcription/Rev+Transcription+Style+Guide+v4.0.1.pdf}}.
\textbf{Filled Pauses (FP)} -- in all our experiments, this group is defined as all \texttt{uh} and \texttt{um} tokens in a given region.
\textbf{Word Fragments (Frag)} -- due to the variability in the format that word fragments can take, we were unable to specify a list of tokens as fragments. Using the ``verbatim" aspect of our transcripts, we define word fragments as all tokens that ended in a ``-".
For each experiment we generate 100,000 samples and report the p-values in \Cref{bias:tab:features}.

\begin{table}[h!]
    \centering
    \begin{tabular}{|c|l|l|l|l|}
        \hline
         \multirow{3}{*}{\textbf{Region}} &  \multicolumn{4}{c|}{\textbf{Feature}}\\
         \cline{2-5}
         & \multicolumn{2}{c|}{\textbf{Before}} & \multicolumn{2}{c|}{\textbf{After}}\\
         \cline{2-5}
         & \multicolumn{1}{c|}{\textbf{FP}} & \multicolumn{1}{c|}{\textbf{Frag}} & \multicolumn{1}{c|}{\textbf{FP}} & \multicolumn{1}{c|}{\textbf{Frag}}\\
         \hline
         African & 0.044$^{*}$&0.026$^{*}$&0.014$^{*}$&0.000$^{**}$\\
         Asian & 0.672&0.000$^{**}$&0.012$^{*}$&0.000$^{**}$\\
         English & 0.003$^{**}$&0.000$^{**}$&0.125&0.000$^{**}$\\
         Germanic & 0.000$^{**}$&0.000$^{**}$&0.770&0.000$^{**}$\\
         Other Romance & 0.000$^{**}$&0.000$^{**}$&0.001$^{**}$&0.000$^{**}$\\
         Slavic & 0.000$^{**}$&0.000$^{**}$&0.003$^{**}$&0.000$^{**}$\\
         Span. / Port. & 0.000$^{**}$&0.000$^{**}$&0.000$^{**}$&0.000$^{**}$\\
         \hline
    \end{tabular}
    \caption{Monte Carlo Permutation Test results comparing the key transcript features in the language regions defined. In the table above, p-values with a $^{*}$ are statistically significant at the 0.05 level while those with $^{**}$ are significant at the 0.005 level.}
    \label{bias:tab:features}
\end{table}

\subsubsection{Results and Discussion}
Our findings in \Cref{bias:tab:features} seemingly contrast and confirm the work in \cite{goldwater2010words}. Our model's ability to recognize tokens before and after a word fragment is impacted by its occurrence. But in contrast to previous work, we find that filled pauses \textit{do} impact our model's recognition in general. Interestingly enough, for the English and Germanic region where the model performs best, we see that our model is capable of recognizing subsequent words well but not preceding words. We also see that the inverse is true of the Asian region indicating higher performance words that come before the filled pause. Though we leave further investigation of these differences to future work, we theorize that the different behaviors over regions occurs due to distinct filled pauses\cite{kosmala:halshs-03225622} used in each region causing model confusion.

\section{Conclusion}
Using \textit{\datasetname} we've shown that, despite major WER improvements on test suites as a whole, the gaps in performance on accented speech in the wild still leaves more to be desired. Using statistical testing, we further highlight that this discrepancy isn't due to random noise but rather a \textit{real underlying} problem of ASR models. 
With the release of this new corpus, we hope to motivate researchers to work on the problem of real-world accented audio. We challenge all industry and academic leaders to find new techniques to improve model recognition on all voices to create more equitable and fair speech technologies.

\section{Acknowledgments}
We would like to thank the transcriptionists without whose help and hundreds of hours of work, none of this dataset release would be possible.

\bibliographystyle{IEEEtran}
\bibliography{main}

\end{document}